%% file: iclr2027_conference.tex
\PassOptionsToPackage{table,xcdraw}{xcolor}
\documentclass{arxiv_preprint} 

\input{math_commands.tex}

\usepackage{hyperref}
\usepackage{url}
\usepackage{booktabs}
\usepackage{amsfonts}       
\usepackage{nicefrac}       
\usepackage{microtype}      
\usepackage{xcolor}         
\usepackage{graphicx}
\usepackage{wrapfig}
\usepackage{amsmath}
\usepackage{algorithm}
\usepackage{algpseudocode}
\usepackage{cleveref}
\usepackage{amssymb}
\usepackage{float}
\usepackage{multirow}
\usepackage{pifont}
\usepackage{wrapfig}
\usepackage{caption}
\usepackage{tabularx}
\usepackage[table]{xcolor}
\newcommand{\cmark}{\ding{51}}
\newcommand{\xmark}{\ding{55}}

\hypersetup{
    pdftitle={On-Policy Parameter Update Direction Underlies Generalization in LLM Post-Training},
    pdfauthor={TODO: authors},
}

\lstdefinestyle{pseudocode}{
    language=Python,
    basicstyle=\ttfamily\small,
    keywordstyle=\color{blue!70!black}\bfseries,
    commentstyle=\color{green!45!black}\itshape,
    stringstyle=\color{orange!70!black},
    numbers=left,
    numberstyle=\tiny\color{gray},
    numbersep=8pt,
    frame=single,
    rulecolor=\color{black!30},
    backgroundcolor=\color{gray!4},
    breaklines=true,
    breakatwhitespace=true,
    showstringspaces=false,
    columns=fullflexible,
    keepspaces=true,
    tabsize=4,
    captionpos=b
}

\newcommand{\corrauth}{\text{\raisebox{-0.12ex}{\scalebox{0.78}{\faEnvelope}}}}

\title{LOOP: Recovery-Aligned Iterative Layer\\ Pruning via On-Policy Distillation}

\author[1,2]{Liang Xu}
\author[1,2]{~Shufan Shen}
\author[3,\corrauth]{~Zhongni Hou}
\author[3]{~Yufei Zhang}
\nextauthorline
\author[3]{Wei Lin} 
\author[3]{~Guojun Yin}
\author[1,\corrauth]{~Shuhui Wang}

\affiliation[1]{State Key Lab. of AI Safety, Institute of Computing Technology, Chinese Academy of Sciences}
\nextaffiliationline
\affiliation[2]{University of Chinese Academy of Sciences}
\affiliation[3]{Meituan}

\contribution[\corrauth]{Corresponding authors}

\metadata[\raisebox{-0.18ex}{\scalebox{0.89}{\faEnvelope}}~Contact]{\email{xuliang23z@ict.ac.cn}, \email{wangshuhui@ict.ac.cn}}

\begin{document}

\input{Section/abstract}
\maketitle
\input{Section/1_intro}
\input{Section/2_relatedwork}
\input{Section/3_preliminaries}
\input{Section/4_method}
\input{Section/5_experiments}
\input{Section/6_conclusion}

\bibliography{iclr2027_conference}
\bibliographystyle{iclr2027_conference}

\newpage
\input{Section/appendix}

\end{document}

%% file: math_commands.tex
\usepackage{amsmath,amsfonts,bm}

\def\eqref#1{equation~\ref{#1}}

\def\1{\bm{1}}

\DeclareMathAlphabet{\mathsfit}{\encodingdefault}{\sfdefault}{m}{sl}
\SetMathAlphabet{\mathsfit}{bold}{\encodingdefault}{\sfdefault}{bx}{n}



%% file: Section/abstract.tex
\begin{abstract}
Layer-level pruning is attractive for large language model (LLM) deployment as it reduces the model size while preserving a dense architecture that can be efficiently executed on standard hardware.
Current methods typically follow a one-shot paradigm, which first removes selected layers to reach a target model size and then performs additional training to recover model capabilities.
However, pruning directly to the target size can substantially degrade reasoning capabilities, making subsequent recovery more difficult under a limited training budget.
Moreover, this paradigm separates layer selection from recovery, limiting the use of recovery information to guide selection and potentially hindering the recovery of reasoning capabilities.
To address these limitations, we propose LOOP, a \textbf{L}ayer-level Recovery-\textbf{O}riented \textbf{O}n-Policy \textbf{P}runing framework that interleaves pruning and recovery across multiple shots. 
In each shot, LOOP removes a single layer and performs short on-policy distillation, mitigating the abrupt degradation in reasoning capabilities associated with one-shot pruning.
To guide layer selection in subsequent shots, LOOP reuses on-policy trajectories and optimization signals from recovery to jointly assess each remaining layer's reasoning contribution and recovery sensitivity, thereby making layer selection recovery-aware.
Evaluations on several benchmarks show that LOOP consistently outperforms one-shot baselines in average accuracy across different model scales and pruning ratios.

\end{abstract}

%% file: Section/1_intro.tex
\section{Introduction}

Large language models (LLMs) \citep{kaplan2020scaling,openai2024,singh2026gpt5} have achieved remarkable capabilities through scaling. However, their increasing memory consumption and computational cost create substantial challenges for efficient deployment.
These deployment challenges have motivated extensive research into model compression, including quantization~\citep{yan2026pt}, knowledge distillation~\citep{shin2026amid}, low-rank decomposition~\citep{zhou2026large,xu2026adaptive}, and pruning~\citep{yang2026entropy}. These approaches reduce memory and computational requirements from different perspectives.

Among these approaches, layer-level pruning removes entire Transformer layers, producing a smaller, shallower model while preserving the remaining network's dense structure~\citep {zhu2024survey}. The resulting model can be efficiently executed on standard hardware using conventional dense computation libraries~\citep{kim2024shortened,yang2024laco}. Together, these properties make layer-level pruning a practical approach for reducing inference cost and memory usage while requiring only minimal changes to the existing model architecture and deployment stack.

Current layer-level pruning methods usually follow a one-shot paradigm that separates compression into two sequential stages: pruning and recovery. 
In this paradigm, layers are typically selected based on importance scores estimated on a calibration prompt set, for example, using changes in representations~\citep{men2025shortgpt}.
The selected layers are then removed together to reach the target model size.
In the recovery stage, the pruned model undergoes additional training to recover reasoning performance~\citep{muralidharan2024compact}.
Despite its practical advantages, layer-level pruning can introduce errors that propagate through subsequent steps of multi-step reasoning.
The one-shot paradigm poses two key challenges to preserving such reasoning capabilities.
First, pruning directly to the target size can substantially degrade reasoning performance and induce a large distributional shift in model predictions. As a result, subsequent recovery starts from a more degraded state, making it harder to restore the lost reasoning capabilities.
Second, one-shot pruning finalizes layer selection before recovery begins. As a result, the importance scores used for pruning may not fully reflect each layer's role in the recovery process.
In particular, scores computed on calibration prompts may fail to capture a layer's contribution along the complete on-policy trajectories encountered during recovery training.
Moreover, criteria that measure only the immediate effect of removing a layer do not directly reflect its influence on the recovery objective. Such criteria may therefore overlook layers that are important for effective recovery.
These limitations highlight the need to limit reasoning degradation during pruning while accounting for each layer's role in subsequent recovery.

To address these limitations, we propose \textbf{LOOP}, a \textbf{L}ayer-level Recovery-\textbf{O}riented \textbf{O}n-Policy \textbf{P}runing framework that interleaves pruning and recovery over multiple shots to preserve reasoning capabilities during compression.
First, to mitigate the severe capability degradation caused by pruning directly to the target size, LOOP adopts an iterative pruning-and-recovery process that distributes layer removal across multiple shots.
In each shot, LOOP removes a single layer and performs a short phase of on-policy recovery. The recovered model serves as the starting point for the next shot, where the remaining layers are re-evaluated before the next pruning decision. This process continues until the target model size is reached. By allowing the model to adapt after each layer removal, LOOP mitigates the abrupt degradation in reasoning capabilities associated with one-shot pruning and supports recovery throughout compression.
Second, to better reflect each layer's role in subsequent recovery, LOOP combines two complementary importance signals: reasoning contribution and recovery sensitivity. Reasoning contribution captures a layer's importance to the reasoning process, while recovery sensitivity reflects its influence on the recovery objective.
For reasoning contribution, LOOP measures the representational changes induced by each remaining layer. The measurement is performed over complete on-policy trajectories generated by the current model during recovery.
For recovery sensitivity, LOOP uses gradient information produced during recovery training. The gradients are used to estimate how strongly the recovery loss depends on each remaining layer, thereby reflecting its influence on the recovery objective.
These two signals are combined into a recovery-oriented importance score, which is updated after each recovery stage to guide layer selection.
Together, these designs enable progressive pruning and recovery while incorporating both reasoning contribution and recovery sensitivity into layer selection.

Extensive experiments on mathematical and scientific reasoning benchmarks demonstrate the effectiveness of LOOP across different pruning ratios and model scales. On Qwen3-4B-Instruct, LOOP outperforms the strongest evaluated one-shot baseline by 7.37, 9.10, and 6.34 percentage points in average mathematical reasoning accuracy at pruning ratios of 15\%, 20\%, and 25\%, respectively.
At a 20\% layer pruning ratio, LOOP consistently outperforms the baselines on Qwen3 models ranging from 1.7B to 8B parameters. LOOP also achieves higher average accuracy than the baseline on scientific reasoning benchmarks across all evaluated pruning ratios, suggesting that the effectiveness of LOOP extends beyond mathematical reasoning.
Further analyses show that both iterative pruning and recovery-oriented layer selection contribute to the performance gains of LOOP.

%% file: Section/2_relatedwork.tex
\section{Related Work}
\subsection{Model Compression and Structured Pruning}
The rapid scaling of large language models~\citep{singh2026gpt5,yang2025qwen3} has led to substantial memory footprints and computational demands, making efficient deployment increasingly challenging. This challenge has motivated a broad range of model compression techniques, including quantization~\citep{park2022lut,liu2025spinquant}, knowledge distillation~\citep{jiang2023lion}, low-rank decomposition~\citep{wang2025dobi,wang2025svd}, and pruning~\citep{ma2023llm,muralidharan2024compact}.
These techniques reduce deployment overhead through different mechanisms: quantization lowers numerical precision, knowledge distillation transfers model capabilities to a compact student, and low-rank decomposition replaces large parameter matrices with compact factors. Pruning, in contrast, reduces model size by removing redundant weights or architectural components from a pretrained network.
Existing pruning methods are commonly categorized as unstructured or structured according to the structure of the sparsity they introduce.
Unstructured pruning sets individual weights to zero using criteria such as weight--activation statistics or local loss sensitivity~\citep{frantar2023sparseGPT,sun2024wanda}, while preserving the original tensor shapes.
However, translating irregular sparsity into memory savings typically requires sparse storage formats, while inference speedups depend on efficient sparse kernels and suitable hardware support~\citep{Liu2023DejaVC}.

In contrast, structured pruning removes entire architectural components, such as attention heads, feed-forward neurons, hidden dimensions, or Transformer layers, producing smaller dense networks that are compatible with standard computation libraries.
Among these granularities, layer-level pruning reduces model depth while preserving the dense structure of the remaining network, making it attractive for efficient deployment~\citep{kim2024shortened}.
Existing structured pruning methods can be further categorized according to how they assess the importance of candidate structures~\citep{zhu2024survey}. 
Loss-based methods estimate structural importance using the loss or its derivatives. For example, LLM-Pruner~\citep{ma2023llm} uses gradient information to identify and remove less important coupled structures. At the layer level, Shortened LLaMA~\citep{kim2024shortened} removes entire Transformer layers and compares LoRA-based fine-tuning with continued pretraining for post-pruning recovery, highlighting the importance of recovery training at high pruning ratios.
Other methods use activation or representation statistics to guide compression. FLAP~\citep{an2023flap} uses input-feature fluctuations to assess the importance of weight-matrix columns and compensates for their removal through bias adjustments. SliceGPT~\citep{ashkboos2024slicegpt} exploits computational invariance to rotate representations into principal-component bases before removing dimensions by deleting corresponding rows and columns of weight matrices. 
At the layer level, ShortGPT~\citep{men2025shortgpt} measures representational changes between layer inputs and outputs using Block Influence (BI) and removes layers with low scores computed on calibration data.
Despite differences in their importance criteria, methods that finalize structural selection before recovery cannot use subsequent recovery signals to revise those decisions. This separation leaves room for approaches that update layer importance during recovery, allowing subsequent pruning decisions to account for both the model's evolving behavior and the recovery objective.

\subsection{Post-Pruning Recovery and On-Policy Distillation}
Since structural removal often degrades model performance, post-pruning recovery is commonly performed through supervised fine-tuning~\citep{ma2023llm}, continued training~\citep{kim2024shortened}, or knowledge distillation~\citep{wang2026layer}. 
Knowledge distillation transfers teacher knowledge to the compressed student through token-level output distributions or teacher-generated responses. The former provides supervision over alternative next tokens, while the latter supplies sequences that the student learns to reproduce~\citep{kim2016sequence}.
Conventional off-policy distillation trains on fixed reference or teacher-generated sequences, whose prefixes may differ from those encountered when the student generates responses at inference time.
On-policy distillation addresses this mismatch by sampling trajectories from the current student and obtaining teacher feedback at the prefixes visited along these trajectories~\citep{agarwal2024OPD,shen2026onpolicy}. As the student is updated, newly generated trajectories allow supervision to follow its evolving behavior.
Recent work has applied on-policy distillation to post-pruning recovery. ShortOPD~\citep{zhang2026shortopd} recovers one-shot depth-pruned models using on-policy distillation with a short-to-long rollout schedule. However, layer selection is completed before recovery, preventing recovery information from informing subsequent structural decisions. This separation makes it difficult to adapt layer selection to the model's behavior during recovery.

%% file: Section/3_preliminaries.tex
\section{Preliminaries}

Let $\mathcal{M}_{pre}$ denote a pretrained language model with $L$ Transformer layers indexed by $\mathcal{I}=\{1,\ldots,L\}$, where  $\mathcal{B}_l$ denotes the transformation of layer $l$, and $\mathcal{E}$ and $\mathcal{H}$ denote the input embedding and output mapping, respectively. The model can be expressed as
\begin{equation}
    \mathcal{M}_{pre}
    =
    \mathcal{H}\circ\mathcal{B}_L\circ\cdots
    \circ\mathcal{B}_1\circ\mathcal{E},
\end{equation}
where $\circ$ denotes function composition.
Given a target layer pruning ratio $\rho$, 
layer-level pruning removes $K=\lceil \rho L\rceil$ layers, where $1\leq K<L$, and retains $L-K$ layers indexed by the retained set $\mathcal{I}_{\mathrm{keep}}=\{i_1,\ldots,i_{L-K}\}\subseteq\mathcal{I}$, with $i_1<\cdots<i_{L-K}$.
The resulting pruned model is
\begin{equation}
\label{math:layer_level_pruning}
    \mathcal{M}_{pruned}
    =
    \mathcal{H}\circ\mathcal{B}_{i_{L-K}}\circ\cdots
    \circ\mathcal{B}_{i_1}\circ\mathcal{E}.
\end{equation}

The retained layers preserve their original order and are initialized with their pretrained parameters.
Recovery training adapts the retained model parameters to mitigate pruning-induced degradation in reasoning performance. The task is to select the retained layers and adapt the model within a prescribed recovery budget, minimizing the degradation in reasoning performance relative to the pretrained model.

%% file: Section/4_method.tex
\section{Method}
\label{method}
\input{Table/Figure_Main}
In this section, we introduce {LOOP}, a layer-level recovery-oriented on-policy pruning framework. As illustrated in~\Cref{fig:method}, LOOP iteratively removes one layer at a time and performs short on-policy recovery through OPD before the next pruning step, allowing the model to adapt progressively throughout compression. During recovery, LOOP reuses on-policy trajectories and recovery signals to re-evaluate the remaining layers, coupling subsequent layer selection with the recovery process.


\subsection{Recovery-Oriented Layer Importance}
Selecting a layer for removal requires considering not only its contribution to the current model's reasoning behavior, but also its relevance to the recovery objective. LOOP captures these two aspects using complementary signals collected during on-policy recovery. Specifically, {reasoning contribution} measures the representational change induced by a layer along on-policy reasoning trajectories, while {recovery sensitivity} measures the local sensitivity of the recovery objective to changes in the layer's contribution. The two signals are aggregated over recovery updates and combined into a recovery-oriented importance score to guide subsequent layer selection.

\paragraph{Reasoning Contribution.}
Reasoning contribution captures how important each layer is to the current student's reasoning process.
LOOP estimates it using Block Influence~(BI), which measures the change between a Transformer layer's input and output representations~\citep{men2025shortgpt}. Whereas conventional BI is typically evaluated on a fixed calibration set, LOOP computes it along complete on-policy trajectories generated by the current student during recovery. This makes the score more representative of the current student's reasoning behavior.
Specifically, consider the recovery stage of shot $k$, which consists of $b_k$ recovery updates. At update $u$, let $p_{\theta}$ denote the current student policy with parameters $\theta$. For a prompt $\mathbf{x}\sim\mathcal{D}$, where $\mathcal{D}$ is the training prompt distribution, the student generates a response 
\begin{equation}
\label{math:on_policy_rollout}
    \mathbf{y}=(y_1,\ldots,y_T)
    \sim p_{\theta}(\cdot\mid\mathbf{x}),
\end{equation}
where $T$ is the response length. The prompt and response are concatenated into a complete trajectory $\boldsymbol{\tau}=[\mathbf{x};\mathbf{y}]$. 
For each remaining layer $l$ at update $u$, let $\mathbf{h}_{\mathrm{in}}^{l}(\boldsymbol{\tau};u,t)$ and $\mathbf{h}_{\mathrm{out}}^{l}(\boldsymbol{\tau};u,t)$ denote its input and output hidden states at token position $t$ in trajectory $\boldsymbol{\tau}$, respectively. The reasoning contribution score at update $u$ is defined as:
\begin{equation}
\label{math:BI}
    s_{\mathrm{Reason}}^{l}(u)
    =
    \mathbb{E}_{\boldsymbol{\tau},t}
    \left[
        1-
        \cos\left(
            \mathbf{h}_{\mathrm{in}}^{l}(\boldsymbol{\tau};u,t),
            \mathbf{h}_{\mathrm{out}}^{l}(\boldsymbol{\tau};u,t)
        \right)
    \right],
\end{equation}
where the expectation denotes an average over all valid tokens in the trajectory batch collected at update $u$, including both prompt and response tokens. Generated reasoning traces are typically longer than prompts in the collected trajectories and thus receive greater aggregate weight. To reduce fluctuations from stochastic rollouts and the influence of isolated score spikes, the reasoning contribution scores are aggregated within each shot using an exponential moving average (EMA,~\cite{tarvainen2017mean}):
\begin{equation}
\label{math:ema_bi}
    \bar{s}_{\mathrm{Reason}}^{l}(u)
    =
    \beta \bar{s}_{\mathrm{Reason}}^{l}(u-1)
    +
    (1-\beta)s_{\mathrm{Reason}}^{l}(u),\qquad u=2,\ldots,b_k,
\end{equation}
where $\beta\in[0,1)$ is the smoothing coefficient. The EMA is initialized as $\bar{s}_{\mathrm{Reason}}^{l}(1)=s_{\mathrm{Reason}}^{l}(1)$ at the first recovery update of each shot.
After $b_k$ steps of updates in shot $k$, the aggregated score for subsequent layer selection is
\begin{equation}
    S_{\mathrm{Reason}}^{l}(k+1)
    =
    \bar{s}_{\mathrm{Reason}}^{l}(b_k).
\end{equation}
A smaller aggregated score indicates less representational change along the sampled on-policy trajectories under the EMA weighting, suggesting a lower reasoning contribution as measured by BI.

\paragraph{Recovery Sensitivity.}
Recovery sensitivity~(RS) assesses each layer's influence on the recovery objective by measuring the sensitivity of the recovery loss to layer gates. 
LOOP estimates RS using OPD Sensitivity, based on gate gradients of the on-policy distillation loss.
During recovery, the student receives supervision from a frozen teacher $\mathcal{T}$, initialized as a copy of $\mathcal{M}_{pre}$, through on-policy distillation~\citep{agarwal2024OPD}. For each response sampled according to \Cref{math:on_policy_rollout}, the teacher additionally conditions on the associated ground-truth answer $a^\star$ and provides a next-token distribution at each prefix $(\mathbf{x},\mathbf{y}_{<t})$. These distributions supervise the student through the following expected token-averaged reverse KL objective:
\begin{equation}
\label{eq:recovery_target}
\mathcal{L}_{\mathrm{Rec}}(\theta)
=
\mathbb{E}_{
    \mathbf{x}\sim\mathcal{D},\,
    \mathbf{y}\sim p_{\theta}(\cdot\mid\mathbf{x})
}
\left[
    \frac{1}{T}
    \sum_{t=1}^{T}
    D_{\mathrm{KL}}
    \left(
        p_{\theta}(\cdot\mid\mathbf{x},\mathbf{y}_{<t})
        \,\middle\|\,
        p_{\mathcal{T}}(\cdot\mid\mathbf{x},a^\star,\mathbf{y}_{<t})
    \right)
\right].
\end{equation}

To measure recovery sensitivity, a scalar gate $g_l$ is introduced for each remaining layer $l$.
For an input representation $\mathbf{h}_{\mathrm{in}}^{l}$ , the gated output is
\begin{equation}
\label{math:layer_gate}
    \widetilde{\mathbf{h}}_{\mathrm{out}}^{l}
    =
    \mathbf{h}_{\mathrm{in}}^{l}
    +
    g_l
    \left[
        \mathcal{B}_l(\mathbf{h}_{\mathrm{in}}^{l})
        -
        \mathbf{h}_{\mathrm{in}}^{l}
    \right].
\end{equation}
Setting $g_l=1$ reproduces the layer output, whereas $g_l=0$ bypasses the layer. During OPD, all gates remain at $1$ and participate in gradient computation but are excluded from optimizer updates.
Let $\mathcal{L}_{\mathrm{Rec}}(u)$ denote the recovery loss on the trajectory batch at update $u$. The recovery sensitivity score of each layer is defined as
\begin{equation}
\label{math:opd_sensitivity}
    s_{\mathrm{Rec}}^{l}(u)
    =
    \left(
        \frac{
            \partial \mathcal{L}_{\mathrm{Rec}}(u)
        }{
            \partial g_l
        }
    \right)^2,
\end{equation}
where the sampled trajectories are held fixed when computing the gate gradients. Squaring the gradient yields a nonnegative measure of local sensitivity, irrespective of the gradient's sign. We use this quantity to measure the local dependence of the recovery objective on each layer.
The recovery sensitivity scores are aggregated within each shot using the same EMA rule as in \Cref{math:ema_bi}. After $b_k$ recovery updates in shot $k$, the aggregated score is
\begin{equation}
    S_{\mathrm{Rec}}^{l}(k+1)
    =
    \bar{s}_{\mathrm{Rec}}^{l}(b_k).
\end{equation}
A larger aggregated score indicates greater local sensitivity of the recovery objective to changes in the layer's contribution, providing a complementary signal for subsequent layer selection.

\paragraph{Combined Importance.}
The aggregated reasoning contribution and recovery sensitivity scores from shot $k$ are combined to guide layer selection in the next shot. To account for differences in their numerical scales, the two sets of EMA-aggregated scores are independently normalized across the remaining layers. For $q\in\{\mathrm{Reason},\mathrm{Rec}\}$, the normalized score is
\begin{equation}
\label{math:norm}
    \widehat{S}_{q}^{l}(k+1)
    =
    \frac{
        S_{q}^{l}(k+1)-\min_j S_{q}^{j}(k+1)
    }{
        \max_j S_{q}^{j}(k+1)-\min_j S_{q}^{j}(k+1)+\epsilon
    },
\end{equation}
where the minimum and maximum are taken over all remaining layers
$j\in\mathcal{I}^{(k+1)}$, and $\epsilon>0$ is a small constant for numerical stability.
The recovery-oriented importance score is then defined as
\begin{equation}
\label{math:combined_importance}
    I_l^{(k+1)}
    =
    \lambda\widehat{S}_{\mathrm{Reason}}^{l}(k+1)
    +
    (1-\lambda)\widehat{S}_{\mathrm{Rec}}^{l}(k+1),
\end{equation}
where $\lambda\in[0,1]$ balances reasoning contribution and recovery sensitivity. A lower $I_l^{(k+1)}$ gives layer $l$ higher priority for removal in the next shot.

\subsection{Iterative Pruning and Recovery}
To mitigate the degradation caused by pruning directly to the target size and to continuously align layer selection with the evolving pruned model, LOOP adopts an iterative pruning and recovery procedure. The procedure consists of three stages: warm-up, iterative pruning and recovery, and final recovery. Together, these stages allow recovery information to guide layer selection throughout compression while further adapting the target-size model after pruning.
The complete procedure is summarized in Appendix~\ref{app:implementation}.

\paragraph{Warm-Up.}
Before the first pruning shot, LOOP performs $b_{\mathrm{w}}$ recovery updates on $\mathcal{M}_{pre}$ without removing any layers. During these updates, on-policy trajectories and gradients of the recovery loss with respect to layer gates are used to compute reasoning contribution and recovery sensitivity, respectively. Both scores are aggregated using EMA, initialized from the first warm-up update. After $b_{\mathrm{w}}$ updates, the resulting model is denoted as $\mathcal{M}^{(0)}$. The aggregated scores are then normalized and combined according to \Cref{math:norm,math:combined_importance} to obtain $I_l^{(0)}$ for the first layer selection.

\paragraph{Iterative Pruning and Recovery.}
At the beginning of shot $k\in\{0,\ldots,K-1\}$, LOOP selects a remaining layer with the lowest combined importance:
\begin{equation}
\label{math:layer_selection}
    l_k^*
    \in
    \arg\min_{l\in\mathcal{I}^{(k)}} I_l^{(k)}.
\end{equation}
The selected layer is removed from the current student, yielding
\begin{equation}
\label{math:layer_pruning}
    \widetilde{\mathcal{M}}^{(k)}
    =
    \operatorname{Prune}
    \left(
        \mathcal{M}^{(k)},l_k^*
    \right),
    \qquad
    \mathcal{I}^{(k+1)}
    =
    \mathcal{I}^{(k)}\setminus\{l_k^*\}.
\end{equation}
The pruned student then undergoes $b_k$ recovery updates:
\begin{equation}
\label{math:iterative_recovery}
    \mathcal{M}^{(k+1)}
    =
    \operatorname{Rec}
    \left(
        \widetilde{\mathcal{M}}^{(k)},
        \mathcal{T};
        b_k
    \right).
\end{equation}
Here, $\operatorname{Rec}(\mathcal{M},\mathcal{T};b)$ denotes $b$ updates to student $\mathcal{M}$ using the recovery objective in \Cref{eq:recovery_target} and the frozen teacher $\mathcal{T}$.
During these updates, on-policy trajectories and gradients of the recovery loss with respect to layer gates are reused to compute reasoning contribution and recovery sensitivity for the remaining layers. The EMA statistics are initialized at the first recovery update of each shot and updated over the remaining recovery steps within each shot. The aggregated scores are normalized and combined to obtain $I_l^{(k+1)}$ for the layer selection in the next shot. This reuse of recovery information avoids collecting a separate set of trajectories for importance estimation.
Each shot therefore removes a single layer, performs recovery, and updates layer importance before the next selection. After $K$ shots, the resulting model $\mathcal{M}^{(K)}$ contains $L-K$ layers.

\paragraph{Final Recovery.}
After $K$ layers have been removed, LOOP performs $b_{\mathrm{f}}$ additional recovery updates without further changes to the model architecture:
\begin{equation}
\label{math:final_recovery}
    \mathcal{M}_{\mathrm{final}}
    =
    \operatorname{Rec}
    \left(
        \mathcal{M}^{(K)},
        \mathcal{T};
        b_{\mathrm{f}}
    \right).
\end{equation}
The total recovery budget $B$, measured in recovery updates, includes warm-up, recovery within each shot, and final recovery:
\begin{equation}
\label{math:recovery_budget}
    b_{\mathrm{w}}
    +
    \sum_{k=0}^{K-1}b_k
    +
    b_{\mathrm{f}}
    = B.
\end{equation}
Together, the three stages allow recovery information to guide layer selection throughout pruning and further adapt the target-size model within the prescribed recovery budget.

%% file: Table/Figure_Main.tex
\begin{figure*}[t]
\vspace{-0.3cm}
\begin{center}
\centerline{\includegraphics[width=1\textwidth]{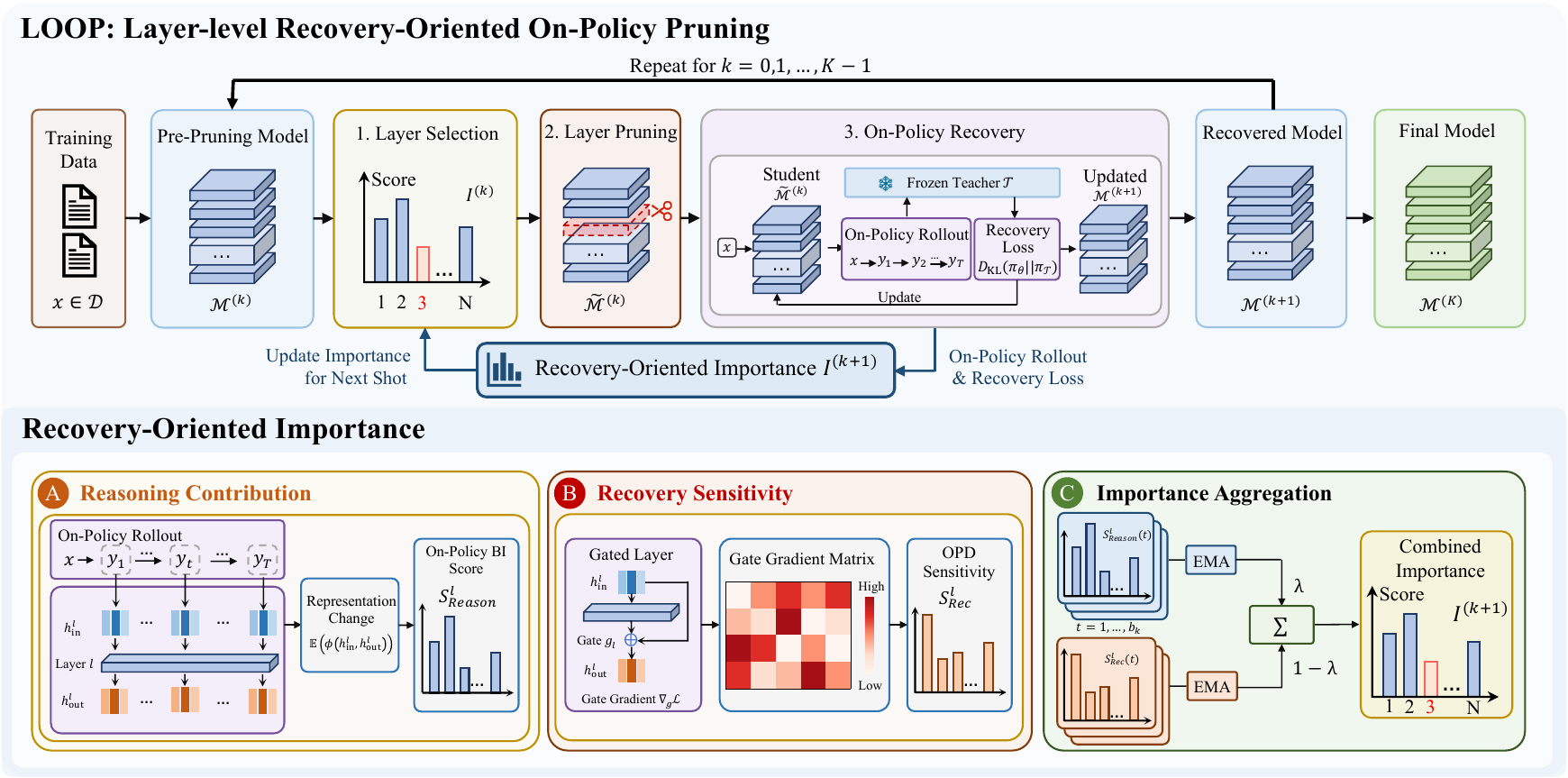}}
\vspace{-0.3cm}
\caption{\textbf{Overview of LOOP.} The top shows the iterative pruning and recovery process, where each shot removes one layer, performs short on-policy recovery, and updates layer importance for the next shot. The bottom details the recovery-oriented importance criterion, which estimates reasoning contribution and recovery sensitivity, aggregates and combines them into an importance score.}
\label{fig:method}
\end{center}
\vspace{-0.6cm}
\end{figure*}

%% file: Section/5_experiments.tex
\section{Experiments}
\label{exp}
\input{Table/Main_Ratio}

We evaluate LOOP on mathematical and scientific reasoning benchmarks across multiple model scales and layer pruning ratios.
Controlled comparisons with one-shot pruning under matched recovery update budgets assess the effects of iterative pruning and recovery-oriented layer selection.
Ablation studies and additional analyses examine the contributions of individual components and the efficiency of the pruning--recovery process.

\subsection{Experimental Setup}

\paragraph{Models and Datasets.}
We evaluate LOOP on three models from the Qwen3 family~\citep{yang2025qwen3}: Qwen3-1.7B, Qwen3-4B-Instruct, and Qwen3-8B. Qwen3-4B-Instruct serves as the primary model for comparisons at target layer pruning ratios of 15\%, 20\%, and 25\%. To assess performance across model scales, all three models are evaluated at a target layer pruning ratio of 20\%.
For mathematical reasoning, models are trained on DAPO-Math-17K~\citep{yu2025dapo} and evaluated on a held-out subset sampled from DAPO-Math, along with MATH-500~\citep{hendrycks2021math}, AMC, and AIME. For scientific reasoning, we use SciKnowEval~\citep{feng2025sciknoweval} for both training and evaluation. The training set combines examples from its Biology, Chemistry, Physics, and Materials domains, while evaluation is conducted separately on the corresponding test sets. Accuracy is used as the evaluation metric for all benchmarks.

\paragraph{Baselines.}
LOOP is compared with baselines using the same pretrained model, training prompts, target layer pruning ratio, and evaluation protocol.
All recovery methods share the same total update budget and common optimization hyperparameters, except for strategy-specific settings.
\textbf{Pruned} removes layers selected by BI~\citep{men2025shortgpt} in a one-shot manner without subsequent recovery.
Starting from the same BI-pruned model, \textbf{SFT w/o KD} performs supervised fine-tuning on fixed ground-truth responses, whereas \textbf{KD} trains on fixed teacher-generated responses.
\textbf{OPD} uses the same BI-pruned initialization but performs on-policy distillation on student-generated responses~\citep{agarwal2024OPD}.
This baseline follows the one-shot pruning and OPD recovery pipeline of ShortOPD~\citep{zhang2026shortopd}, but uses a fixed maximum rollout length.
The unpruned model \textbf{Original} serves as a reference.

\subsection{Main Experimental Results}

\paragraph{Performance under Different Pruning Ratios.}
To assess the effectiveness of LOOP at different levels of pruning, \Cref{tab:main_qwen3_4b} compares mathematical reasoning performance on Qwen3-4B-Instruct at target layer pruning ratios of 15\%, 20\%, and 25\%.
Pruning without recovery causes severe performance degradation, demonstrating that directly removing entire Transformer layers substantially disrupts the model's capability.
Under matched recovery update budgets, LOOP achieves the highest accuracy among the evaluated methods across all four benchmarks and pruning ratios. Compared with the best baseline, LOOP improves the average accuracy by 7.37, 9.10, and 6.34 percentage points at pruning ratios of 15\%, 20\%, and 25\%, respectively. It also retains 83.40\%, 60.46\%, and 45.52\% of the original model's average performance at the corresponding ratios.
These results demonstrate the advantage of LOOP in preserving reasoning performance across different ratios.

\input{Table/Main_Model}
\input{Table/Main_Science}

\paragraph{Performance across Model Scales.}
The cross-scale comparison evaluates whether LOOP's effectiveness extends to models of different sizes.
\Cref{tab:main_different_scales} reports results on Qwen3-1.7B, Qwen3-4B-Instruct, and Qwen3-8B at a target layer pruning ratio of 20\%.
LOOP achieves the highest average accuracy at each evaluated scale, outperforming the strongest baseline at each scale by 1.08, 9.10, and 10.84 percentage points on Qwen3-1.7B, Qwen3-4B-Instruct, and Qwen3-8B, respectively.
These results support the effectiveness of LOOP across model scales.

\paragraph{Scientific Reasoning Performance.}
Experiments on the Biology, Chemistry, Physics, and Materials domains of SciKnowEval further assess the effectiveness of LOOP beyond mathematical reasoning.
As shown in \Cref{tab:main_science}, LOOP outperforms the OPD baseline in terms of average accuracy across all pruning ratios. Specifically, LOOP improves the average accuracy over the OPD baseline by 5.00, 3.85, and 2.12 percentage points at pruning ratios of 15\%, 20\%, and 25\%, respectively, while retaining 77.30\%, 67.71\%, and 61.18\% of the original model's average performance.

Across the evaluated settings, LOOP consistently achieves higher average accuracy than the compared baselines. These results support the effectiveness of recovery-oriented multi-shot pruning across pruning ratios, model scales, and reasoning domains under the prescribed recovery budget.

\subsection{Ablation Study and Analysis}
\input{Table/Ablation_Components}
\paragraph{Effect of Individual Components.}
\Cref{tab:ablation_loop} isolates the contributions of iterative pruning and recovery-oriented importance on Qwen3-4B-Instruct at a 20\% layer pruning ratio.
Starting from the one-shot BI baseline at 23.58\% mean accuracy, introducing iterative pruning alone improves accuracy to 30.69\%, while incorporating recovery sensitivity into BI under one-shot pruning raises it to 26.74\%.
Combining both components yields the highest mean accuracy of 32.68\%, exceeding iterative BI by 1.99 percentage points and one-shot pruning with combined importance by 5.94 percentage points.
These results show that each component improves performance both individually and when added to the other, supporting their complementary roles in post-pruning recovery.

\input{Table/Ablation_OPD_BI}
\paragraph{Importance Fusion Strategy.}
To examine how reasoning contribution and recovery sensitivity should be combined for layer selection, \Cref{tab:fusion_strategy} compares fusion strategies on Qwen3-4B-Instruct at a 15\% layer pruning ratio. The two-stage strategy first filters candidate layers using BI and then selects a layer using recovery sensitivity. Weighted fusion combines the two normalized scores according to \Cref{math:combined_importance}, with $\lambda$ controlling the weight of reasoning contribution.
Equal weighting ($\lambda=0.5$) achieves the highest accuracy on all four benchmarks, with an average accuracy of 0.451, compared with 0.329 for two-stage selection and 0.399 for recovery sensitivity alone. Unequal weighting also yields lower average accuracy, indicating that the relative weighting of the two signals matters. These results favor balanced fusion for layer selection in this setting.

\paragraph{Efficiency Analysis.}
\Cref{tab:training_efficiency} compares LOOP with the one-shot OPD baseline in terms of measured training time and response length.
Per-step time is averaged over the full training run and includes the complete wall-clock duration of each training step.
At layer pruning ratios of 20\% and 25\%, the average per-step time decreases from 53.50 to 48.89 seconds and from 54.65 to 49.88 seconds, respectively.
Under matched evaluation settings, the final LOOP models also generate responses that are 12.67\% and 17.51\% shorter, respectively.
With the same retained depth at each pruning ratio, shorter responses require fewer autoregressive decoding steps.
Together, these results show that LOOP improves final-model performance without increasing average training-step time or response length under the evaluated settings.

%% file: Table/Main_Ratio.tex
\begin{table*}[!t]
\caption{Performance of Qwen3-4B-Instruct under different layer-pruning ratios and recovery strategies on mathematical reasoning benchmarks. All results are measured by accuracy. \textit{Mean} denotes the average accuracy across the four benchmarks, and $T$ denotes the compressed model's mean performance relative to that of the original model.}
\label{tab:main_qwen3_4b}
\centering
\resizebox{\textwidth}{!}{
\begin{tabular}{c|l|l|c|ccccc|c}
\toprule
Ratio & Method & Recovery & Metric
& DAPO & Math-500 & AMC & AIME & Mean & $T$ (\%) \\
\midrule

0\%
& Original
& -- & --
& 0.472 & 0.784 & 0.650 & 0.256 & 0.5405 & 100.000 \\
\midrule

\multirow{5}{*}{15\%}
& Pruned
& -- & BI
& 0.032 
& 0.138 
& 0.000 
& 0.000 
& 0.0425 
& 7.865\\

& SFT w/o KD
& SFT & BI
& 0.143  
& 0.464 
& 0.175 
& 0.033  
& 0.2038 
& 37.719 \\

& KD
& Off-Policy & BI
& 0.124 
& 0.478 
& 0.175 
& 0.011 
& 0.1970 
& 36.454 \\

& OPD
& On-Policy & BI
& \underline{0.332}
& \underline{0.604}
& \underline{0.450}
& \underline{0.122}
& \underline{0.3771}
& \underline{69.769} \\

\rowcolor{gray!20}
& \textbf{LOOP (Ours)}
& On-Policy & BI + RS
& \textbf{0.348}
& \textbf{0.658}
& \textbf{0.575}
& \textbf{0.222}
& \textbf{0.4508}
& \textbf{83.404} \\

\midrule

\multirow{5}{*}{20\%}
& Pruned
& -- & BI
& 0.004 
& 0.014 
& 0.000 
& 0.000 
& 0.0045 
& 0.833 \\

& SFT w/o KD
& SFT & BI
& 0.072  
& 0.312 
& 0.050 
& 0.011  
& 0.1113 
& 20.592 \\

& KD
& Off-Policy & BI
& 0.052 
& 0.338 
& 0.075 
& 0.000 
& 0.1163 
& 21.512 \\

& OPD
& On-Policy & BI
& \underline{0.139}
& \underline{0.468}
& \underline{0.325}
& \underline{0.011}
& \underline{0.2358}
& \underline{43.626} \\

\rowcolor{gray!20}
& \textbf{LOOP (Ours)}
& On-Policy & BI + RS
& \textbf{0.238}
& \textbf{0.522}
& \textbf{0.425}
& \textbf{0.122}
& \textbf{0.3268}
& \textbf{60.463} \\

\midrule

\multirow{5}{*}{25\%}
& Pruned
& -- & BI
& 0.001 
& 0.006 
& 0.000 
& 0.000 
& 0.0018 
& 0.324\\

& SFT w/o KD
& SFT & BI
& 0.023  
& 0.190  
& 0.050 
& 0.000 
& 0.0658 
& 12.167 \\

& KD
& Off-Policy & BI
& 0.029 
& 0.202 
& 0.050 
& 0.000 
& 0.0703 
& 13.000 \\

& OPD
& On-Policy & BI
& \underline{0.175}
& \underline{0.386}
& \underline{0.125}
& \underline{0.044}
& \underline{0.1826}
& \underline{33.792} \\

\rowcolor{gray!20}
& \textbf{LOOP (Ours)}
& On-Policy & BI + RS
& \textbf{0.192}
& \textbf{0.470}
& \textbf{0.200}
& \textbf{0.122}
& \textbf{0.2460}
& \textbf{45.522} \\

\bottomrule
\end{tabular}
}
\vspace{-0.3cm}
\end{table*}

%% file: Table/Main_Model.tex
\begin{table*}[!t]
\vspace{-0.2cm}
\caption{Performance of different-sized Qwen3 models under a 20\% layer-pruning ratio on mathematical reasoning benchmarks. All results are measured by accuracy. The best and second-best results among the reported compression methods are highlighted in \textbf{bold} and \underline{underlined}, respectively.}
\label{tab:main_different_scales}
\centering
\resizebox{\textwidth}{!}{
\begin{tabular}{l|l|l|c|ccccc|c}
\toprule
Model & Method & Recovery & Metric
& DAPO & Math-500 & AMC & AIME & Mean & $T$ (\%) \\
\midrule

\multirow{6}{*}{Qwen3-1.7B}
& Original
& -- & --
& 0.063 & 0.242 & 0.150 & 0.011 & 0.1165 & 100.000 \\

\cmidrule(lr){2-10}
& Pruned
& -- & BI
& 0.001  & 0.008  & 0.000  & 0.000  & 0.0023  & 1.931 \\

& SFT w/o KD
& SFT & BI
& \underline{0.031}  
& 0.130
& 0.000 
& 0.000   
& 0.0403  
& 34.542 \\

& KD
& Off-Policy & BI
& 0.028 
& 0.122 
& 0.025 
& 0.011 
& 0.0465 
& 39.929 \\

& OPD
& On-Policy & BI
& 0.011
& \underline{0.154}
& \textbf{0.075}
& \underline{0.000}
& \underline{0.0600}
& \underline{51.491} \\

\rowcolor{gray!20}
& \textbf{LOOP (Ours)}
& On-Policy & BI + RS
& \textbf{0.032}
& \textbf{0.190}
& \underline{0.050}
& \textbf{0.011}
& \textbf{0.0708}
& \textbf{60.717} \\

\midrule

\multirow{6}{*}{Qwen3-4B}
& Original
& -- & --
& 0.472 & 0.784 & 0.650 & 0.256 & 0.5405 & 100.000 \\

\cmidrule(lr){2-10}
& Pruned
& -- & BI
& 0.004 
& 0.014 
& 0.000 
& 0.000 
& 0.0045 
& 0.833 \\

& SFT w/o KD
& SFT & BI
& 0.072  
& 0.312 
& 0.050 
& 0.011  
& 0.1113 
& 20.592 \\

& KD
& Off-Policy & BI
& 0.052 
& 0.338 
& 0.075 
& 0.000 
& 0.1163 
& 21.512 \\

& OPD
& On-Policy & BI
& \underline{0.139}
& \underline{0.468}
& \underline{0.325}
& \underline{0.011}
& \underline{0.2358}
& \underline{43.626} \\

\rowcolor{gray!20}
& \textbf{LOOP (Ours)}
& On-Policy & BI + RS
& \textbf{0.238}
& \textbf{0.522}
& \textbf{0.425}
& \textbf{0.122}
& \textbf{0.3268}
& \textbf{60.463} \\

\midrule

\multirow{6}{*}{Qwen3-8B}
& Original
& -- & --
& 0.197 & 0.414 & 0.325 & 0.056 & 0.2479 & 100.000 \\

\cmidrule(lr){2-10}
& Pruned
& -- & BI
& 0.007  & 0.010  & 0.000  & 0.000  & 0.0043  & 1.714 \\

& SFT w/o KD
& SFT & BI
& \underline{0.107}
& \underline{0.352}
& 0.150
& \underline{0.011}
& \underline{0.1550}
& \underline{62.539} \\

& KD
& Off-Policy & BI
& 0.084 
& 0.260 
& 0.150 
& 0.011 
& 0.1263 
& 50.941 \\

& OPD
& On-Policy & BI
& 0.071
& 0.338
& \underline{0.175}
& \underline{0.011}
& 0.1488
& 60.007 \\

\rowcolor{gray!20}
& \textbf{LOOP (Ours)}
& On-Policy & BI + RS
& \textbf{0.108}
& \textbf{0.440}
& \textbf{0.350}
& \textbf{0.156}
& \textbf{0.2634}
& \textbf{106.247} \\

\bottomrule
\end{tabular}
}
\vspace{-0.2cm}
\end{table*}

%% file: Table/Main_Science.tex
\begin{table*}[!t]
\caption{Performance of Qwen3-4B-Instruct under different layer-pruning ratios on scientific reasoning benchmarks~\citep{feng2025sciknoweval}. All results are measured by accuracy.}
\label{tab:main_science}
\centering
\resizebox{\textwidth}{!}{
\begin{tabular}{c|l|c|ccccc|c}
\toprule
Ratio & Method & Metric
& Biology & Chemistry & Materials & Physics & Mean & $T$ (\%) \\
\midrule

0\%
& Original
& --
& 0.200 & 0.138 & 0.574 & 0.400 & 0.3281 & 100.000 \\
\midrule

\multirow{2}{*}{15\%}
& OPD
& BI
& 0.040 & 0.090 & \textbf{0.521} & 0.163 & 0.2036 & 62.040 \\

& \textbf{LOOP (Ours)}
& BI + RS
& \textbf{0.180} & \textbf{0.126} & \textbf{0.521} & \textbf{0.188} & \textbf{0.2536} & \textbf{77.296} \\
\midrule

\multirow{2}{*}{20\%}
& OPD
& BI
& 0.040 & 0.090 & 0.447 & \textbf{0.158} & 0.1837 & 55.985 \\

& \textbf{LOOP (Ours)}
& BI + RS
& \textbf{0.160} & \textbf{0.100} & \textbf{0.479} & 0.150 & \textbf{0.2222} & \textbf{67.714} \\
\midrule

\multirow{2}{*}{25\%}
& OPD
& BI
& 0.080 & \textbf{0.111} & 0.389 & 0.138 & 0.1795 & 54.704 \\

& \textbf{LOOP (Ours)}
& BI + RS
& \textbf{0.120} & 0.052 & \textbf{0.468} & \textbf{0.163} & \textbf{0.2007} & \textbf{61.180} \\
\bottomrule
\end{tabular}
}
\vspace{-0.3cm}
\end{table*}

%% file: Table/Ablation_Components.tex
\begin{table}[!ht]
\vspace{-0.3cm}
\caption{Ablation study of LOOP using Qwen3-4B-Instruct at a 20\% compression ratio. Combined Importance indicates whether recovery sensitivity is incorporated with BI for layer selection.}
\label{tab:ablation_loop}
\centering
\begin{tabularx}{0.97\textwidth}{
    cc
    >{\centering\arraybackslash}X
    c
    *{3}{>{\centering\arraybackslash}X}
}
\toprule
Iterative Pruning & Combined Importance
& DAPO & Math-500 & AMC & AIME & Mean \\
\midrule
\rowcolor{gray!20}
\cmark & \cmark
& \underline{0.238}
& \textbf{0.522}
& \textbf{0.425}
& \textbf{0.122}
& \textbf{0.3268} \\

\cmark & \xmark
& \textbf{0.261}
& \underline{0.514}
& \underline{0.375}
& \underline{0.078}
& \underline{0.3069} \\

\xmark & \cmark
& 0.185 
& 0.482 
& 0.325 
& \underline{0.078}
& 0.2674  \\

\xmark & \xmark
& 0.139
& 0.468
& 0.325
& 0.011
& 0.2358 \\
\bottomrule
\end{tabularx}

\vspace{-0.2cm}
\end{table}

%% file: Table/Ablation_OPD_BI.tex

\begin{table*}[ht]
\centering

\begin{minipage}[t]{0.569\textwidth}
\vspace{0pt}
\centering
\begin{minipage}[t][2.8\baselineskip][t]{\linewidth}
\captionof{table}{
Comparison of different importance fusion settings for Qwen3-4B-Instruct under a 15\% layer-pruning ratio. All results are reported using accuracy.
}
\label{tab:fusion_strategy}
\end{minipage}

\vspace{0.1cm}
\resizebox{\linewidth}{!}{
\begin{tabular}{llccccc}
\toprule
Strategy & Setting & DAPO & Math-500 & AMC & AIME & Mean \\
\midrule
Two-stage
& BI $\rightarrow$ RS
& 0.258 & 0.510 & 0.425 & 0.122 & 0.329 \\
\midrule
\multirow{3}{*}{Weighted}
& $\lambda=0.8$
& 0.283 & 0.590 & 0.450 & 0.133 & 0.364 \\
& $\lambda=0.5$
& \textbf{0.348}
& \textbf{0.658}
& \textbf{0.575}
& \textbf{0.222}
& \textbf{0.451} \\
& $\lambda=0.2$
& \underline{0.344}
& 0.598
& \underline{0.550}
& \underline{0.156}
& \underline{0.412} \\
\midrule
RS only
& $\lambda=0$
& 0.301
& \underline{0.600}
& \underline{0.550}
& 0.144
& 0.399 \\
\bottomrule
\end{tabular}
}
\end{minipage}
\hfill
\begin{minipage}[t]{0.412\textwidth}
\vspace{0pt}
\centering
\begin{minipage}[t][2.8\baselineskip][t]{\linewidth}
\captionof{table}{
Comparison of the efficiency of LOOP and OPD on Qwen3-4B-Instruct under different pruning ratios.
}
\label{tab:training_efficiency}
\end{minipage}

\vspace{0.1cm}
\resizebox{\linewidth}{!}{
\begin{tabular}{ccc}
\toprule
Method
& Time~(s)
& Output Tokens \\
\midrule
LOOP-20\%
& 48.888
& \textbf{13678.75} \\

OPD-20\%
& 53.496
& 15662.13 \\
\midrule

LOOP-25\%
& 49.878
& \textbf{14014.75} \\

OPD-25\%
& 54.648
& 16990.00 \\
\bottomrule
\end{tabular}
}
\end{minipage}

\vspace{-0.3cm}
\end{table*}

%% file: Section/6_conclusion.tex
\section{Conclusion}
In this work, we presented \textbf{LOOP}, a layer-level recovery-oriented on-policy pruning framework designed to limit reasoning degradation during compression.
LOOP alternates single-layer pruning with short on-policy distillation and uses recovery signals to guide subsequent layer selection.
Its layer-selection criterion combines reasoning contribution, estimated from representational changes along complete on-policy trajectories, with recovery sensitivity, estimated from squared gradients of the recovery loss with respect to layer gates.
Experiments on mathematical and scientific reasoning benchmarks show that LOOP achieves higher average accuracy than the evaluated one-shot baselines under matched recovery update budgets across multiple pruning ratios and model scales.
Ablation studies support the contributions of both iterative pruning and recovery-oriented importance.
These findings highlight the value of accounting for recovery in layer selection to better preserve reasoning performance under a limited recovery budget.

%% file: Section/appendix.tex
\appendix
\section{Interpretation of Recovery Sensitivity}

Recovery sensitivity is designed to characterize how strongly the recovery objective locally depends on the contribution of each layer. Consider the recovery loss as a function of the layer gate $g_l$. Around the operating point $g_l=1$, a first-order Taylor expansion gives
\begin{equation}
\mathcal{L}_{\mathrm{Rec}}(1+\Delta g_l)
\approx
\mathcal{L}_{\mathrm{Rec}}(1)
+
\frac{\partial \mathcal{L}_{\mathrm{Rec}}}{\partial g_l}
\Delta g_l.
\end{equation}
Thus, the magnitude of $\partial\mathcal{L}_{\mathrm{Rec}}/\partial g_l$ reflects how strongly a local change in the contribution of layer $l$ affects the recovery objective. A gradient close to zero indicates weak local dependence, whereas a gradient with large magnitude indicates that the recovery objective is more sensitive to that layer.

Because recovery sensitivity measures the strength of this dependence rather than its direction, LOOP uses the squared gate gradient. For a small zero-mean perturbation $\Delta g_l$ with variance $\sigma^2$,
\begin{equation}
\mathbb{E}\left[(\Delta\mathcal{L}_{\mathrm{Rec}})^2\right]
\approx
\sigma^2
\left(
\frac{\partial\mathcal{L}_{\mathrm{Rec}}}{\partial g_l}
\right)^2.
\end{equation}
This gives the squared gate gradient a direct interpretation as a nonnegative measure of local recovery sensitivity. It also prevents gradients with opposite directions from canceling when sensitivity is aggregated across on-policy recovery updates. Therefore, recovery sensitivity serves as a local recovery-oriented importance signal rather than an estimate of the exact loss change caused by removing a layer.

\section{Algorithm}
\label{app:implementation}
Algorithm~\ref{alg:loop} summarizes the three-stage procedure of LOOP. The \textsc{RecoverAndScore} subroutine is defined in Algorithm~\ref{alg:recover_score}, where on-policy trajectories and gate gradients are reused to estimate layer importance. The operator $\mathrm{Rec}(\mathcal{M},\mathcal{T};b)$ denotes the same OPD recovery updates without importance estimation. Here, $\mathcal{M}_{\mathrm{pre}}$ denotes the pretrained model, while $\mathcal{M}^{(0)}$ denotes the student after warm-up.
\input{Table/Algorithm_LOOP}

\section{Implementation Details}
\label{app:ExpDetail}

\paragraph{Training Configuration.}
The experiments use vLLM for rollout generation.
The training batch size and optimization mini-batch size are both 32, with one response sampled per prompt.
The learning rate is $2\times10^{-5}$, with a learning-rate warm-up of 10 updates.
The total training budget is fixed at 200 update steps.
OPD and LOOP share the same maximum rollout length and total update budget to isolate the effects of iterative pruning and layer selection.

\paragraph{Rollout and Pruning Configuration.}
Training responses are sampled with a temperature of 0.6, top-$p$ of 0.9, and top-$k$ of 50.
Pruning is restricted to complete Transformer layers.
We use $b_w=20$ warm-up updates before the first pruning shot
and reserve a nominal budget of $\bar{b}_f=40$ updates for
final recovery.
For a total budget of $B=200$ updates and $K$ pruning shots,
each shot is assigned
\begin{equation}
    b_k = \left\lceil
    \frac{B-b_w-\bar{b}_f}{K}
    \right\rceil,
    \qquad k=0,\ldots,K-1.
\end{equation}
Any additional updates introduced by rounding up are deducted
from the final recovery budget, giving
\begin{equation}
    b_f = B-b_w-\sum_{k=0}^{K-1} b_k.
\end{equation}
This allocation ensures that the total number of updates
remains exactly $B$.
For LOOP, weighted fusion is enabled with equal weights of 0.5 for reasoning contribution~(measured by BI) and recovery sensitivity.
The EMA decay $\beta$ is set to 0.9.
The one-shot comparison uses BI-based layer selection with an EMA decay of 0.9.

\paragraph{Teacher Configuration.}
During on-policy distillation, the student generates responses using only the original problem prompt.
For teacher inference, the original prompt is augmented with the ground truth using a fixed instruction template.
The student-generated response is then appended to this augmented prompt, and the teacher computes token distributions at the corresponding student-generated positions.

\section{Computational Complexity and Scalability}
For a model with $L$ Transformer layers and target pruning ratio $\rho$, LOOP performs
$K=\lceil \rho L\rceil$
pruning shots. The total number of recovery updates is controlled by the fixed budget
\begin{equation}
b_w+\sum_{k=0}^{K-1}b_k+b_f=B.
\end{equation}
For this depth-based comparison, we assume matched token workloads.
Let $C_{\mathrm{stu}}(m)$ denote the student-side cost of one recovery update when the student contains $m$ layers, including on-policy generation and optimization, and let $C_{\mathrm{tea}}(L)$ denote the teacher inference cost. Since the teacher remains fixed while the student becomes progressively shallower, the total recovery computation is
\begin{equation}
\begin{aligned}
C_{\mathrm{LOOP}}
=&\; b_w C_{\mathrm{stu}}(L)
+\sum_{k=0}^{K-1} b_k C_{\mathrm{stu}}(L-k-1)
+b_f C_{\mathrm{stu}}(L-K) \\
&+B C_{\mathrm{tea}}(L).
\end{aligned}
\end{equation}
For comparison, a one-shot OPD pipeline that first estimates layer importance on a calibration set, prunes directly to $L-K$ layers, and then performs $B$ recovery updates has cost
\begin{equation}
C_{\mathrm{OneShot}}
=
C_{\mathrm{cal}}(L)
+
B C_{\mathrm{stu}}(L-K)
+
B C_{\mathrm{tea}}(L),
\end{equation}
where $C_{\mathrm{cal}}(L)$ denotes the cost of the separate calibration stage. Therefore, the difference in computational cost between the two methods is
\begin{equation}
\begin{aligned}
    C_{\mathrm{LOOP}} - C_{\mathrm{OneShot}} = &b_w \left(C_{\mathrm{stu}}(L) - C_{\mathrm{stu}}(L-K)\right)\\
                            &+\sum_{k=0}^{K-1} b_k \left(C_{\mathrm{stu}}(L-k-1) - C_{\mathrm{stu}}(L-K)\right)\\
                            & - C_{\mathrm{cal}}(L)
\end{aligned}
\end{equation}
Assuming that the student-side computation of each recovery update scales approximately linearly with model depth, we write $C_{\mathrm{stu}}(m)=cm$, where $c$ denotes the average computation per Transformer layer. The difference between LOOP and one-shot OPD can then be expressed as
\begin{equation}
\begin{aligned}
C_{\mathrm{LOOP}}-C_{\mathrm{OneShot}}
=&\; c\left[
b_w K+
\sum_{k=0}^{K-1} b_k(K-k-1)
\right]
-C_{\mathrm{cal}}(L)\\
\approx&
c\left[
b_wK+
\frac{(B-b_w-b_f)(K-1)}{2}
\right]
-C_{\mathrm{cal}}(L).
\end{aligned}
\end{equation}
This analysis highlights the additional student-side computation from retaining more layers during early recovery, offset in part by avoiding separate calibration.
Actual wall-clock time also depends on rollout lengths and implementation details.
Table~\ref{tab:training_efficiency} reports lower average training-step time for LOOP under the evaluated settings.

\section{Experiments}
\subsection{Extended Recovery Analysis}
\label{app:extended_recovery}
\input{Table/Appendix_Loss}
To examine whether the performance gap persists with additional recovery training, we compare LOOP with a one-shot BI baseline on Qwen3-4B-Instruct at a 20\% layer pruning ratio.
The one-shot baseline in these figures corresponds to the OPD baseline in the main experiments, where all target layers are first pruned using BI and the resulting model is then recovered with on-policy distillation. \Cref{fig:extended_accuracy} compares LOOP with this baseline at evaluation checkpoints from 200 to 500 updates, where OOD Avg denotes the mean accuracy on MATH-500, AMC and AIME.
LOOP achieves higher DAPO accuracy and higher average accuracy on the out-of-distribution benchmarks at every reported checkpoint.
Although accuracy fluctuates during training, the one-shot baseline does not close the gap within the evaluated update range.
These results suggest that the advantage of LOOP persists beyond the initial recovery budget.
\Cref{fig:recovery_loss} compares the recovery loss of LOOP and the one-shot BI baseline throughout training.
Although LOOP's loss initially rises with fluctuations, it remains below that of the one-shot baseline throughout the observed training period.
After 200 updates, both losses show an overall downward trend, while the gap persists.
Together with LOOP's consistently higher accuracy at the reported evaluation checkpoints, these results suggest that its pruning strategy facilitates the recovery of reasoning capabilities.

\subsection{Effect of OPD Without Pruning}
\input{Table/Appendix_OPD}

To disentangle the effect of on-policy distillation from that of pruning and recovery, we further apply OPD directly to the unpruned Qwen3-8B model under the same training budget.
As shown in Table~\ref{tab:app_Qwen_8B}, OPD improves the mean accuracy of the unpruned model from 0.2479 to 0.3548, indicating that on-policy training itself provides substantial task adaptation.
Nevertheless, the pruned models remain considerably more challenging to recover: the one-shot OPD baseline reaches only 0.1488 mean accuracy, while LOOP achieves 0.2634.
Importantly, both methods use on-policy distillation for recovery, yet LOOP outperforms the one-shot OPD baseline by 11.46 percentage points.
This comparison suggests that the gains of LOOP over one-shot pruning cannot be attributed to on-policy distillation alone, and instead support the benefit of coupling iterative pruning with recovery-oriented layer selection.

\subsection{Cross-Domain Generalization}
\input{Table/Appendix_cross_domain}

To assess cross-domain performance retention after recovery on mathematical data, we evaluate models pruned and recovered on DAPO-Math on scientific reasoning benchmarks without additional training on scientific data.
As shown in Table~\ref{tab:cross_domain_science}, LOOP achieves a mean accuracy of 0.0951, compared with 0.0285 for the one-shot OPD baseline, with improvements on Materials and Physics.
These results suggest that LOOP mitigates cross-domain performance degradation relative to one-shot OPD.

\subsection{Results on Llama-3.1-8B.}
\input{Table/Appendix_Llama}
To assess whether the benefits of LOOP extend beyond the Qwen3 family, we additionally evaluate it on Llama-3.1-8B~\citep{grattafiori2024llama3} at a 20\% layer pruning ratio.
As shown in \Cref{tab:llama_results}, LOOP outperforms the one-shot OPD baseline on all four benchmarks, improving mean accuracy from 7.73\% to 11.30\%.
These results provide additional evidence of LOOP's effectiveness beyond Qwen3, although performance remains below that of the unpruned model.

%% file: Table/Algorithm_LOOP.tex
\begin{algorithm}[ht]
\caption{LOOP: Layer-level Recovery-Oriented On-Policy Pruning}
\label{alg:loop}
\begin{algorithmic}[1]
\Require Pretrained model $\mathcal{M}_{\mathrm{pre}}$ with $L$ layers;
prompt distribution $\mathcal{D}$; pruning ratio $\rho$;
update budgets $b_{\mathrm{w}},\{b_k\}_{k=0}^{K-1},b_{\mathrm{f}}$;
EMA coefficient $\beta$; fusion weight $\lambda$;
normalization constant $\epsilon>0$.
\Ensure Compressed model $\mathcal{M}_{\mathrm{final}}$.

\Statex
\State $K \gets \lceil \rho L\rceil$, with $1\leq K<L$
\State Require $b_{\mathrm{w}}\geq1$, $b_k\geq1$, $b_{\mathrm{f}}\geq0$,
and $b_{\mathrm{w}}+\sum_{k=0}^{K-1}b_k+b_{\mathrm{f}}=B$
\State $\mathcal{T}\gets\operatorname{Copy}(\mathcal{M}_{\mathrm{pre}})$;
freeze $\mathcal{T}$
\State $\mathcal{M}\gets\operatorname{Copy}(\mathcal{M}_{\mathrm{pre}})$
\State $\mathcal{I}^{(0)}\gets\{1,\ldots,L\}$

\Statex
\State \textbf{Stage 1: Warm-up}
\State $(\mathcal{M}^{(0)},\mathbf{I}^{(0)})\gets
\Call{RecoverAndScore}{
    \mathcal{M},\mathcal{T},\mathcal{D},
    \mathcal{I}^{(0)},b_{\mathrm{w}},\beta,\lambda,\epsilon
}$

\Statex
\State \textbf{Stage 2: Iterative pruning and recovery}
\For{$k=0,\ldots,K-1$}
    \State Select $l_k^*\in
    \arg\min_{l\in\mathcal{I}^{(k)}}I_l^{(k)}$
    using a fixed tie-breaking rule
    \State $\widetilde{\mathcal{M}}^{(k)}
    \gets\operatorname{Prune}(\mathcal{M}^{(k)},l_k^*)$
    \State $\mathcal{I}^{(k+1)}
    \gets\mathcal{I}^{(k)}\setminus\{l_k^*\}$
    \State $(\mathcal{M}^{(k+1)},\mathbf{I}^{(k+1)})\gets
    \Call{RecoverAndScore}{
        \widetilde{\mathcal{M}}^{(k)},\mathcal{T},\mathcal{D},
        \mathcal{I}^{(k+1)},b_k,\beta,\lambda,\epsilon
    }$
\EndFor

\Statex
\State \textbf{Stage 3: Final recovery}
\State $\mathcal{M}_{\mathrm{final}}\gets
\operatorname{Rec}(\mathcal{M}^{(K)},\mathcal{T};b_{\mathrm{f}})$
\State \Return $\mathcal{M}_{\mathrm{final}}$
\end{algorithmic}
\end{algorithm}

\begin{algorithm}[ht]
\caption{\textsc{RecoverAndScore}: On-Policy Recovery with Importance Estimation}
\label{alg:recover_score}
\begin{algorithmic}[1]

\Procedure{RecoverAndScore}
{$\mathcal{M},\mathcal{T},\mathcal{D},\mathcal{I},b,\beta,\lambda,\epsilon$}

\State Freeze $\mathcal{T}$; attach differentiable gates
$\mathbf{g}=\mathbf{1}$ excluded from optimizer updates
\For{$u=1,\ldots,b$}
    \State Sample $(\mathbf{x},a^\star)\sim\mathcal{D}$ and generate
    $\mathbf{y}\sim p_\theta(\cdot\mid\mathbf{x})$;
    form $\boldsymbol{\tau}=[\mathbf{x};\mathbf{y}]$
    \State Compute $\mathcal{L}_{\mathrm{Rec}}(u)$
    using the teacher conditioned on $a^\star$
    \State Obtain the parameter and gate gradients
    of $\mathcal{L}_{\mathrm{Rec}}(u)$
    \State Compute $\{s_{\mathrm{Reason}}^l(u),
    s_{\mathrm{Rec}}^l(u)\}_{l\in\mathcal{I}}$
    using \Cref{math:BI} and~\Cref{math:opd_sensitivity}
    \If{$u=1$}
        \State $\bar{s}_q^l\gets s_q^l(u)$
        for all $l\in\mathcal{I}$,
        $q\in\{\mathrm{Reason},\mathrm{Rec}\}$
    \Else
        \State $\bar{s}_q^l\gets
        \beta\bar{s}_q^l+(1-\beta)s_q^l(u)$
        for all $l,q$
    \EndIf
    \State Update $\mathcal{M}$ using the OPD gradients;
    keep $\mathbf{g}=\mathbf{1}$
\EndFor
\State $S_q^l\gets\bar{s}_q^l$; normalize each score type
using \Cref{math:norm}
\State $I_l\gets
\lambda S_{\mathrm{Reason}}^l+
(1-\lambda)S_{\mathrm{Rec}}^l$
for all $l\in\mathcal{I}$
\State Remove auxiliary gates
\State \Return $(\mathcal{M},\mathbf{I})$

\EndProcedure
\end{algorithmic}
\end{algorithm}

%% file: Table/Appendix_Loss.tex
\begin{figure}[ht]
    \centering
    \begin{minipage}[t]{0.49\linewidth}
        \vspace{0pt}
        \centering
        \includegraphics[width=1\textwidth]{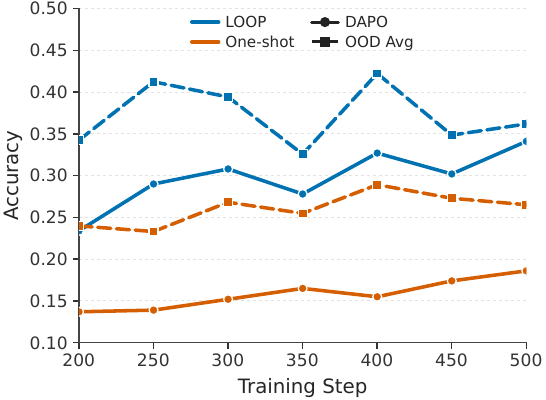}
        \caption{Accuracy of LOOP and the one-shot BI baseline
        during extended recovery training.}
        \label{fig:extended_accuracy}
    \end{minipage}
    \hfill
    \begin{minipage}[t]{0.49\linewidth}
        \vspace{0pt}
        \centering
        \includegraphics[width=1\textwidth]{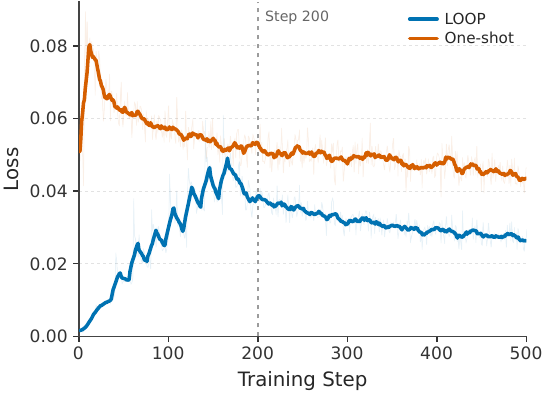}
        \caption{Recovery loss of LOOP and the one-shot BI
        baseline over 500 updates.}
        \label{fig:recovery_loss}
    \end{minipage}
\end{figure}

%% file: Table/Appendix_OPD.tex
\begin{table*}[!t]
\vspace{-0.2cm}
\caption{Comparison of OPD training without pruning and pruning-based recovery methods on Qwen3-8B. Pruned methods use a 20\% layer-pruning ratio.}
\label{tab:app_Qwen_8B}
\centering
\resizebox{\textwidth}{!}{
\begin{tabular}{l|l|l|c|ccccc|c}
\toprule
Model & Method & Recovery & Metric
& DAPO & Math-500 & AMC & AIME & Mean & $T$ (\%) \\
\midrule

\multirow{6}{*}{Qwen3-8B}
& Original
& -- & --
& 0.197 & 0.414 & 0.325 & 0.056 & 0.2479 & 100.00 \\

& OPD Trained
& -- & --
& 0.273  & 0.510  & 0.425  & 0.211  & 0.3548  & 143.120 \\

\cmidrule(lr){2-10}
& Pruned
& -- & BI
& 0.007  & 0.010  & 0.000  & 0.000  & 0.0043  & 1.71 \\

& OPD
& On-Policy & BI
& 0.071
& 0.338
& \underline{0.175}
& \underline{0.011}
& 0.1488
& 60.01 \\

\rowcolor{gray!20}
& \textbf{LOOP (Ours)}
& On-Policy & BI + RS
& \textbf{0.108}
& \textbf{0.440}
& \textbf{0.350}
& \textbf{0.156}
& \textbf{0.2634}
& \textbf{106.25} \\

\bottomrule
\end{tabular}
}
\vspace{-0.2cm}
\end{table*}

%% file: Table/Appendix_cross_domain.tex
\begin{table}[!ht]
\centering
\caption{Cross-domain scientific reasoning performance of Qwen3-4B-Instruct at a 20\% layer-pruning ratio. 
All results are reported using accuracy.}
\label{tab:cross_domain_science}
\resizebox{\columnwidth}{!}{
\begin{tabular}{cccccccccc}
\toprule
Ratio & Method & Recovery Data & Metric 
& Biology & Chemistry & Materials & Physics & Mean & $T$ (\%) \\
\midrule

0\% 
& Original 
& -- 
& -- 
& 0.200 
& 0.138 
& 0.574 
& 0.400 
& 0.3281 
& 100.000 \\

\midrule

\multirow{4}{*}{20\%}
& OPD
& Science
& BI
& 0.040
& 0.090
& 0.447
& \textbf{0.158}
& 0.1837
& 55.985 \\

& LOOP (Ours)
& Science
& BI + RS
& \textbf{0.160}
& \textbf{0.100}
& \textbf{0.479}
& 0.150
& \textbf{0.2222}
& \textbf{67.714} \\

& OPD
& Math
& BI
& 0.000
& 0.000
& 0.064
& 0.050
& 0.0285
& 8.673 \\

& LOOP (Ours)
& Math
& BI + RS
& 0.000
& 0.000
& 0.255
& 0.125
& 0.0951
& 28.977 \\

\bottomrule
\end{tabular}
}
\end{table}

%% file: Table/Appendix_Llama.tex
\begin{table}[ht]
\centering
\caption{Performance of Llama-3.1-8B on mathematical reasoning
benchmarks. All results are measured by accuracy.}
\label{tab:llama_results}
\resizebox{\columnwidth}{!}{
\begin{tabular}{llcrrrrrr}
\toprule
 Method & Recovery & Metric
& DAPO & Math-500 & AMC & AIME & Mean & $T$ (\%) \\
\midrule
Original & -- & --
& 0.160 & 0.452 & 0.275 & 0.022 & 0.2273 & 100.000 \\
\midrule

OPD & On-Policy & BI
& 0.015 & 0.194 & 0.100 & 0.000 & 0.0773 & 33.985 \\
LOOP (Ours) & On-Policy & BI + RS
& \textbf{0.097} & \textbf{0.200} & \textbf{0.150}
& \textbf{0.005} & \textbf{0.1130} & \textbf{49.713} \\
\bottomrule
\end{tabular}
}
\end{table}